\documentclass[runningheads]{llncs}
\usepackage{graphicx}

\usepackage{tikz}
\usepackage{comment}
\usepackage{amsmath,amssymb} 
\usepackage{color}

\usepackage[accsupp]{axessibility}  

\usepackage{hyperref}
\usepackage[inline]{enumitem}
\usepackage{ntheorem}

\makeatletter
\newcommand{\thickhline}{%
    \noalign {\ifnum 0=`}\fi \hrule height 1pt
    \futurelet \reserved@a \@xhline
}
\makeatother

\usepackage{caption} 

\makeatletter
\renewcommand{\paragraph}{%
  \@startsection{paragraph}{4}%
  {\z@}{0.5ex}{-1em}%
  {\normalfont\normalsize\bfseries}%
}
\makeatother

\usepackage{xcolor}
\definecolor{bittersweet}{rgb}{1.0, 0.44, 0.37}
\definecolor{mygreen}{rgb}{0.29, 0.7, 0.48}

\begin{document}
\pagestyle{headings}
\mainmatter
\def\ECCVSubNumber{3214}  

\title{Object-Centric Unsupervised Image Captioning} 

\titlerunning{Object-Centric Unsupervised Image Captioning}
%
\author{Zihang Meng\inst{1} \and
David Yang\inst{2} \and Xuefei Cao\inst{2} \and Ashish Shah\inst{2} \and Ser-Nam Lim\inst{2}}
\authorrunning{Z. Meng et al.}
%
\institute{$^1$ University of Wisconsin-Madison     $^2$ Meta AI\\
\email{zihangm@cs.wisc.edu; xuefeicao01@gmail.com; dzyang,ashishbshah,sernamlim@fb.com}
}

\onecolumn{%
\renewcommand\twocolumn[1][]{#1}%
\maketitle
\begin{center}
    \centering
    \captionsetup{type=figure}
    \includegraphics[width=0.92\textwidth]{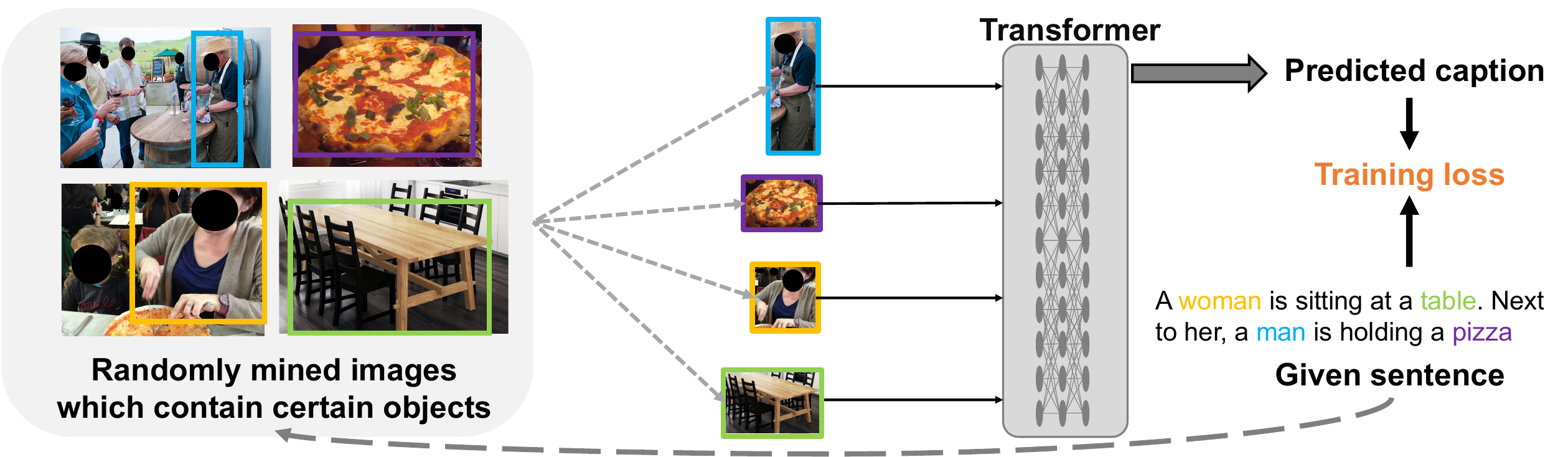}
    \captionof{figure}{Overview of our image captioning training pipeline. During training, the transformer takes the set of object regions mined from the entire image dataset as input. During test/inference, it takes in the object regions of the given image.}
    \label{fig:overall_arch}
\end{center}
}



\begin{abstract}

Image captioning is a longstanding problem in the field of computer vision and natural language processing. To date, researchers have produced impressive state-of-the-art performance in the age of deep learning. Most of these state-of-the-art, however, requires large volume of annotated image-caption pairs in order to train their models. When given an image dataset of interests, practitioner needs to annotate the caption for each image in the training set and this process needs to happen for each newly collected image dataset. In this paper, we explore the task of unsupervised image captioning which utilizes unpaired images and texts to train the model so that the texts can come from different sources than the images. A main school of research on this topic that has been shown to be effective is to construct pairs from the images and texts in the training set according to their overlap of objects. Unlike in the supervised setting, these constructed pairings are however not guaranteed to have fully overlapping set of objects. Our work in this paper overcomes this by harvesting objects corresponding to a given sentence from the training set, even if they don't belong to the same image. When used as input to a transformer, such mixture of objects enables larger if not full object coverage, and when supervised by the corresponding sentence, produced results that outperform current state of the art unsupervised methods by a significant margin. Building upon this finding, we further show that (1) additional information on relationship between objects and attributes of objects also helps in boosting performance; and (2) our method also extends well to non-English image captioning, which usually suffers from a scarcer level of annotations. Our findings are supported by strong empirical results. Our code is available at \href{https://github.com/zihangm/obj-centric-unsup-caption}{https://github.com/zihangm/obj-centric-unsup-caption}.
\end{abstract}

\section{Introduction}
\label{sec:intro}


Image captioning is an important task standing at the crossroad of computer vision (CV) and natural language processing (NLP) that has been widely studied for many years. In the deep learning era, with the advent of transformer models~\cite{zhou2020unified,changpinyo2019decoupled,meng2021connecting}, significant advances in image captioning have been made since its ``humble'' beginning from the early use of Convolutional Neural Networks in combination with Recurrent Neural Networks \cite{karpathy2015deep}. Since then, various attention mechanisms \cite{xu2015show,anderson2018bottom} and transformer based models \cite{cornia2020meshed} have been proposed with great effect. 
The current success has however been predicated on the availability of large amount of image-caption annotations, which is quite expensive to obtain. As a matter of fact, in \cite{pont2020connecting}, it was revealed that it costs 144.7 seconds on average for a professional full-time annotator to provide a high-quality caption for just a single image. 

This has led researchers to propose methods that do not require image-caption pairings, but instead train their models on separate image and text datasets. This line of work precipitates the onset of unsupervised image captioning, with
\cite{feng2019unsupervised} making an early attempt here by utilizing policy gradient to encourage visual concepts in the predicted captions. This approach, however, only encourages the appearance of visual object words, but ignores how they should properly fit into the sentence. Later on, \cite{laina2019towards} proposed to mine pseudo image-caption pairs to train the model. Given a sentence, the algorithm searches and pairs the sentence with an image in the training set which contains overlapping visual concepts. Building on this, \cite{honda2021removing} makes further improvements by introducing a new gate function that tells the model which word in the sentence is irrelevant to the image to form higher quality pseudo image-caption pairs. 

While these advances have produced state of the art performance, we found that they have a fundamental limitation. Since the image and text datasets are unpaired, it is more likely that the quality of the image-caption pairs could be sub par as measured by the number of objects in the sentence that are actually captured in the image (see Table~\ref{tab:gcc_ss_ablation} and \ref{tab:honda_ss_ablation} on how object coverage affects performance). Our work in this paper tries to solve this problem with a simple yet effective approach. Given a sentence in the text dataset, instead of trying to find a candidate from the image dataset, we harvest objects corresponding to the sentence. Because we do not require these objects to be from the same image, we significantly increase the chance of fully covering all the objects that appear in the sentence. Specifically, the harvested objects are fed into a transformer, which is then supervised by the corresponding sentence during training. ``Surprisingly'', experiments show that our approach outperforms the state-of-the-art methods by a clear margin.  

To further boost performance, we note that the harvested objects do not really respect spatial relationships (e.g., the phrase ``person riding a bike'' requires the ``person'' to be above ``bike''). However, when a relationship detector also becomes available, our approach naturally enables the utilization of such relationship information by feeding such information together with the objects into the transformer. Most previous works \cite{feng2019unsupervised,laina2019towards,honda2021removing,guo2020recurrent} would find the incorporation of such information challenging without making significant change to their model or training procedure.

Finally, we explore the possibility of going beyond English to generate captions in other languages. Non-English captioning tasks are expected to be one of the largest benefactors of unsupervised image captioning, simply because paired image-captions annotations are scarcely available in languages other than English, and really speaks to the importance of making advances in unsupervised image caption. We demonstrate our proposed approach on non-English captioning with convincing empirical results.
\section{Related Work}
\textbf{Supervised Image Captioning.}
Supervised image captioning traditionally relies on paired image-caption data to train a generative model which creates a text description given an input image. In recent years, the research community has significantly raised the level of performance for the image captioning task \cite{stefanini2021show}. Some earlier work such as \cite{karpathy2015deep} adopts Convolutional Neural Networks (CNNs) and Recurrent Neural Networks (RNNs) with global image feature as input, while others such as \cite{xu2015show,bahdanau2014neural} proposed to add attention over the grid of CNN features. \cite{anderson2018bottom} further adds attention over visual regions to learn better feature representations. More recently, transformer-based models have been utilized with great success \cite{cornia2020meshed}. Other notable advances include personalized image captioning \cite{zhang2020learning} that is conditioned on the learned representation of a certain user, dense image captioning \cite{johnson2016densecap} which localizes and describes salient image regions, and the generation of captions in a controllable way guided by speech \cite{deshpande2019fast}, a set of bounding boxes \cite{cornia2019show} or human attention traces \cite{meng2021connecting}. {One recent work called CLIP \cite{radford2021learning} proposes  to learn visual models using noisy image-caption pairs automatically mined from the internet. The model is not designed for captioning task but can possibly be used as a pre-trained visual and language embedding model for the captioning task.}

\textbf{Towards Unsupervised Image Captioning.}
Paired image-caption datasets are usually very expensive to obtain. To deal with this challenge, some studies have been conducted to explore the possibility of reducing the amount of paired information needed or utilizing unpaired images and sentences. \cite{venugopalan2017captioning,hendricks2016deep} explored learning simultaneously from
multiple data sources with auxiliary objectives to describe a variety of objects unseen in paired image-caption data. \cite{gu2018unpaired} leverages the paired information in a pivot language to train the generative model and translate the generated captions into the target language. More relevant to our work is the recent line of work in \cite{feng2019unsupervised,laina2019towards,honda2021removing,liu2021exploring,guo2020recurrent,gu2019unpaired}, which utilize unpaired image and text information. They assume that a pretrained object detector is available for extracting visual concepts from an image which act as the link between visual and language domain. Some of them utilize adversarial training to align the visual and language domain. \cite{feng2019unsupervised} relies on discrete rewards to help the model generate high-quality captions. \cite{guo2020recurrent} trains a generator which generates a sentence from discrete words. They discard the visual information in detected object regions and only keep the visual concept words as the input to the model. \cite{gu2019unpaired} assumes that a pretrained scene graph detector is available, and trains a captioning model using scene graph decomposition. \cite{laina2019towards} mines pseudo image-caption pairs from existing but unpaired image and text datasets to train the captioning model. \cite{honda2021removing} improves on it by removing spurious alignments. Our work in this paper follows the same line of thoughts to mine pseudo image-caption pairs. However, unlike these existing approaches which when given a sentence look for \emph{an} image that contains as many of the objects occurring in the sentence as possible, our method explores the possibility of harvesting these objects from different images. 

\textbf{Non-English Captioning.}
Most current works on image captioning focus on the English language. To extend captioning technology to non-English languages, we are starting to see some studies being reported. Some researchers have attempted to directly propose a captioning model on a target language while utilizing a pivot language, typically English, in which paired information is readily available \cite{gu2018unpaired,lan2017fluency,song2019unpaired,wu2019improving}. Nevertheless, the straightforward approach remains to collect image-caption pairs in the target language (e.g., French \cite{rajendran2015bridge}, German \cite{elliott2016multi30k}, or Chinese \cite{li2019coco}). These efforts have not been as extensive as hoped, and advances in unsupervised image captioning could have strong impact here. In this paper, we provide benchmarks showing that even in non-English captioning tasks, our proposed method surpasses or is on par with the state of the art unsupervised image captioning methods.

\section{Method}
\label{sec:method}
Given a set of images $\mathcal{I}=\{I_1,..., I_{N_i}\}$, and a set of sentences $\mathcal{S}=\{S_1,..., S_{N_s}\}$, our goal is to train a model which takes an image as input and generates a caption that well describes the input image. We follow previous work on unsupervised image captioning \cite{feng2019unsupervised} in assuming that we do not have information about the pairing between $\mathcal{I}$ and $\mathcal{S}$, but have access to a pretrained object detector (a pretrained Faster R-CNN \cite{anderson2018bottom}). Our overall approach is shown in Fig. \ref{fig:overall_arch}, where we propose to mine objects from $\mathcal{I}$ without the constraint that these objects need to come from the same image. Specifically, given a sentence $S_i$, we first detect the set of visual concepts it contains (e.g., person, bike) as
\begin{equation}
    \mathcal{V}^S_i = \{v_1, v_2, ..., v_k\},
\end{equation}
where $k$ is the number of visual concepts in the sentence $S_i$. From $\mathcal{V}^S_i$, we mine a set of image regions (we use the pretrained Faster R-CNN to detect the regions and extract the visual features) given as
\begin{equation}
\label{eq:img_region_objs}
    \mathcal{R}_i = \{r_1,r_2,...,r_k\},
\end{equation}
where each  $r_j\in \mathcal{R}_i$ is the visual feature of the object region of visual concept $v_j\in \mathcal{V}^S_i$ taken from a randomly chosen image $I \in \mathcal{I}$ which contains this visual concept. We further encode the location information by concatenating the 5-D vector representing the bounding box locations (four coordinates and the area, from the image $I$ where the region is mined) with $r_j$. $\mathcal{R}_i$ forms the training pair with the given sentence $S_i$. We can utilize these artificially constructed $\mathcal{R}$--$S$ pairs to train the captioning model using Cross Entropy (CE) loss:
\begin{equation}
    L = \text{CE}(f(\mathcal{R}_i), S_i),
\end{equation}
where $f$ represents the captioning model which predicts the caption from a set of object region features. In this work, we adopt the transformer model described in \cite{vaswani2017attention} as $f$.



\subsection{Utilizing Additional Information}
\label{sec:utilize_more_info}
The flexibility of a transformer architecture means that we can also include additional information as input together with the object information. In our method described so far, we note that a part of the information is lost, being that we used the original locations of the object regions, which may be incorrect in relation to other objects as well as losing fine-grained information when we replaced the objects (e.g., face attributes). In this regard, our experiments also show that losing such information causes a drop in performance (see Table~\ref{tab:verify_hyp1}). To this end, we describe how we take a pretrained relationship detector and attribute detector as examples of additional information that we can add to our method.

Consider that for a given $I_i$, the relationship detector can detect triplets in the form of subject-relation-object,
\begin{equation}
    \mathcal{T}^I_i = \{\text{sub}_1\text{-rel}_1\text{-obj}_1, ..., \text{sub}_{N_i^I}\text{-rel}_{N_i^I}\text{-obj}_{N_i^I}\},
\end{equation}
and for a sentence $S_i$, the relationship parser can detect triplets in the same form,
\begin{equation}
    \mathcal{T}^S_i = \{\text{sub}_1\text{-rel}_1\text{-obj}_1, ..., \text{sub}_{N_i^S}\text{-rel}_{N_i^S}\text{-obj}_{N_i^S}\}.
\end{equation}
Then for a given sentence $S_i$, we can similarly obtain $\mathcal{R}_i$ using the same strategy as that in Eqn. \ref{eq:img_region_objs}. However, instead of only selecting image regions which match the visual concepts in $S_i$, we also add to $\mathcal{R}_i$ pairs of image regions (sub-obj) which match the triplets (sub-rel-obj) in $\mathcal{T}^S_i$:
\begin{equation}
    \mathcal{R}_i = \{r_1, ..., r_k, pair_1, ..., pair_l\},
\end{equation}
where each $r$ refers to one image region and each $pair$ refers to a pair of image regions formed by the subject region and the object region from one triplet. $k$ is the number of visual concepts in $S_i$ and $l$ is the number of triplets in $S_i$ detected by the language relationship detector. The subsequent steps to construct pairs for training are the same as that in Sec. \ref{sec:method}.

To utilize the pretrained attribute detector, we basically perform the same steps. The only difference is that instead of detecting triplets subject-relation-object, the attribute detector detects pairs in the form of attribute-object. Then when we see a certain attribute-object pair in the sentence $S_i$, we look for an image region that matches this attribute-object, and add this image region into $\mathcal{R}_i$. Similar procedure could be conducted when other pretrained models such as scene detector, facial expression detector, etc., are available. In this paper, we will focus on leveraging relationship and attribute detectors in addition to the objects, and leave the study of adding other detectors to future work.

\subsection{Extension to Other Languages}
\label{sec:method_non_english}
We can also easily adapt our work to train a non-English image captioning model in an unsupervised way. Consider the same object detector pretrained on English vocabulary. We denote the English vocabulary of the object detector as
\begin{equation}
    \text{voc}_\text{English} = \{\text{word}_1, ..., \text{word}_{N_\text{English}}\}.
\end{equation}
Then we translate each word in the English vocabulary into the target language denoted as ``Target'' (although translating a whole sentence into a different language is a challenging task, translating a single word can be easily done using language dictionaries) given as
\begin{equation}
    \text{voc}_\text{Target} = \{\text{word}_1, ..., \text{word}_{N_\text{Target}}\},
\end{equation}
where $N_\text{English}$ may not equal to $N_\text{Target}$ considering it is possible that multiple words in English are translated into the same word in the target language and vice versa. After this, we can follow the same steps in Sec. \ref{sec:method} to construct pairs to train a model directly in the target language.

\section{Experiments}
Our experiments are designed to empirically establish:
\begin{enumerate*}[label=(\roman*)]
    \item that our proposed method outperforms state of the art unsupervised image captioning methods,
    \item that our proposed method can easily incorporate additional information, in this case object relationships and attributes to help boost performance, 
    \item the effects of swapping in objects from different images to generate training pairs, and,
    \item that our proposed method outperforms state of the art in non-English captioning tasks.
\end{enumerate*}
We will provide the implementation details next so that we have a common ground for discussing the results.
\\

{\noindent\textbf{Collecting Text Corpus from the Internet.} One key advantage of unsupervised image captioning is being able to utilize text corpus directly mined from the internet.  In the experiments, we use two internet-mined text corpus collected by previous works: Shutterstock (SS) \cite{feng2019unsupervised} and Google Conceptual Captions (GCC) \cite{sharma2018conceptual}. SS is a set of sentences collected from the  Shutterstock website using the filter that each sentence should have some object words overlapped with COCO object categories. The sentences in GCC  are automatically harvested from the Alt-text HTML attribute associated with web images. For some cases (described in later sections), we also utilize sentences from existing human annotated image caption datasets, mainly to demonstrate the property of our proposed method.}
\\

\noindent\textbf{Implementation Details.}
We tokenize the text datasets and delete sentences which are shorter than $5$ words or longer than a certain length, 20 for GCC and SS, and 100 for Localized Narratives \cite{pont2020connecting}) and build a language vocabulary. Then we match the object words of the object detector with the language vocabulary. 
Our object detector is a Faster R-CNN \cite{ren2015faster} pretrained on Visual Genome \cite{krishna2017visual}, whose vocabulary contains 1600 object words. Next, we build a dictionary that maps an object word to a set of images which contain this object word. Finally, during training, for a randomly picked sentence, we first find all object words it contains, and for each object word, we randomly pick one image that contains this object word and crop this object region using its bounding box location. In this way, for each sentence $S_i$ we can have a set of object regions  $R_i$. Our captioning model is a one layer transformer \cite{vaswani2017attention}. The size of the hidden attention layers is 512 and that of the feed-forward layers is 2048. The input object features are extracted by the Faster R-CNN mentioned above. We train the network with a batch size of $100$ using the Adam optimizer \cite{kingma2014adam}. The initial learning rate is 5e-4, which decays by 0.8 every 3 epochs, for a total of 30 epochs. The same training setup is used for all experiments in this paper. {\color{black} Note that these hyperparameters are directly borrowed from \cite{meng2021connecting} (except that we increased the batch size from 30 to 100 for faster training), and we did not utilize the validation set to further finetune the hyperparameters to ensure that we do not utilize any pairing information during training (the validation set contains pairing information). }
\\

\noindent\textbf{Evaluation Datasets.}
For images, we follow previous work \cite{feng2019unsupervised,honda2021removing} to use the MS COCO dataset and the train/validation/test split provided by \cite{karpathy2015deep}, and report the performance on the test split (except in Sec. \ref{sec:verify_hyp}, where we follow \cite{meng2021connecting} to use COCO-2017 official splits). For the text, we choose the recently released Localized Narratives (LN) \cite{pont2020connecting} which provides  captions for four public datasets including COCO, ADE20k, Flickr30k and Open Images. We choose LN-COCO instead of COCO captions \cite{chen2015microsoft} because the captions in LN are longer, contain more verbs, and the overall quality is better 
(see our supplement or Table 2 in \cite{pont2020connecting} for a comparison between LN-COCO and COCO captions). We use LN-COCO for all experiments except those on non-English languages in Sec \ref{sec:non-eng}, where we use the annotations provided by COCO-CN \cite{li2019coco} and Multi30k \cite{elliott2016multi30k}.
\\

\noindent\textbf{Evaluating Metrics.} 
We use the official COCO caption evaluation tool and report the performance in terms of BLEU-1 \cite{papineni2002bleu}, BLEU-4, METEOR \cite{banerjee2005meteor}, ROUGE \cite{lin2004rouge}, CIDEr \cite{vedantam2015cider}, SPICE \cite{anderson2016spice} and WMD \cite{kusner2015word}.

\subsection{Comparisons With State-of-the-Art Methods}
\label{sec:GCC_SS}

\begin{table*}[!h]
    \centering
    \begin{tabular}{l l | c c c c c c c }
        \thickhline
        Text dataset & Method & BLEU-1 & BLEU-4 & METEOR & ROUGE$_L$ & CIDEr & SPICE & WMD \\
        \hline 
        SS & \cite{feng2019unsupervised} & 0.016 & 0.001 & 0.037 & 0.109 & 0.018 & 0.073 & 0.045 \\
        SS & \cite{honda2021removing} & 0.022 & 0.001 & 0.043 & 0.126 & 0.025 & 0.078 & 0.042 \\
        SS & Ours & \textbf{0.056} & \textbf{0.003} & \textbf{0.060} & \textbf{0.127} & \textbf{0.038} & \textbf{0.102} & \textbf{0.060}\\
        \hline
        GCC & \cite{honda2021removing} & 0.006 & 0.000 & 0.035 & 0.115 & 0.017 & 0.075 & 0.040 \\
        GCC & Ours & \textbf{0.062} & \textbf{0.004} & \textbf{0.062} & \textbf{0.146} & \textbf{0.032} & \textbf{0.104} & \textbf{0.055} \\
        \thickhline
    \end{tabular}
    \caption{The performance of our method and baseline methods \cite{feng2019unsupervised,honda2021removing} trained using COCO images and SS/GCC text datasets, evaluated using test split of COCO images and LN-COCO caption annotations as ground truth. }
    \label{tab:gcc_ss}
\end{table*}

\begin{table*}[!h]
    \centering
    \resizebox{1.01\columnwidth}{!}{
    \begin{tabular}{l | c c c c c c c }
        \thickhline
        Pretrained models & BLEU-1 & BLEU-4 & METEOR & ROUGE$_L$ & CIDEr & SPICE & WMD \\
        \hline 
        object  & 0.327 & 0.059 & 0.140 & 0.262 & 0.109 & 0.181 & 0.079 \\
        object + attribute & \textbf{0.332} & 0.059 & 0.136 & 0.266 & 0.124 & 0.181 & 0.079\\
        object + relationship  & 0.329 & 0.061 & 0.140 & 0.268 & 0.120 & 0.188 & 0.080 \\
        object + relationship + attribute & 0.329 & \textbf{0.062} & \textbf{0.141} & \textbf{0.274} & \textbf{0.138} & \textbf{0.193} & \textbf{0.083}\\
        \thickhline
    \end{tabular}
    }
    \caption{The performance of utilizing only object detector and that of utilizing object detector plus relationship/attribute detector. The models are trained using COCO images and LN-OpenImages captions, and evaluated using the test split of COCO images and LN-COCO caption annotations as ground truth.}
    \label{tab:rel}
\end{table*}

\cite{feng2019unsupervised,honda2021removing} are two state of the art methods in unsupervised image captioning. We compare our method with them using COCO images together with GCC and SS as the text datasets respectively. We use the code released by the authors to produce results on these datasets for benchmarking. It is important to note that both GCC and SS are web-crawled text datasets and help to showcase unsupervised methods' flexibility in exploiting large scale data by breaking the chain of pairing. We further note that these state of the art methods are also not transformer based.

\begin{figure*}[!h]
    \centering
    \includegraphics[width=1.01\linewidth]{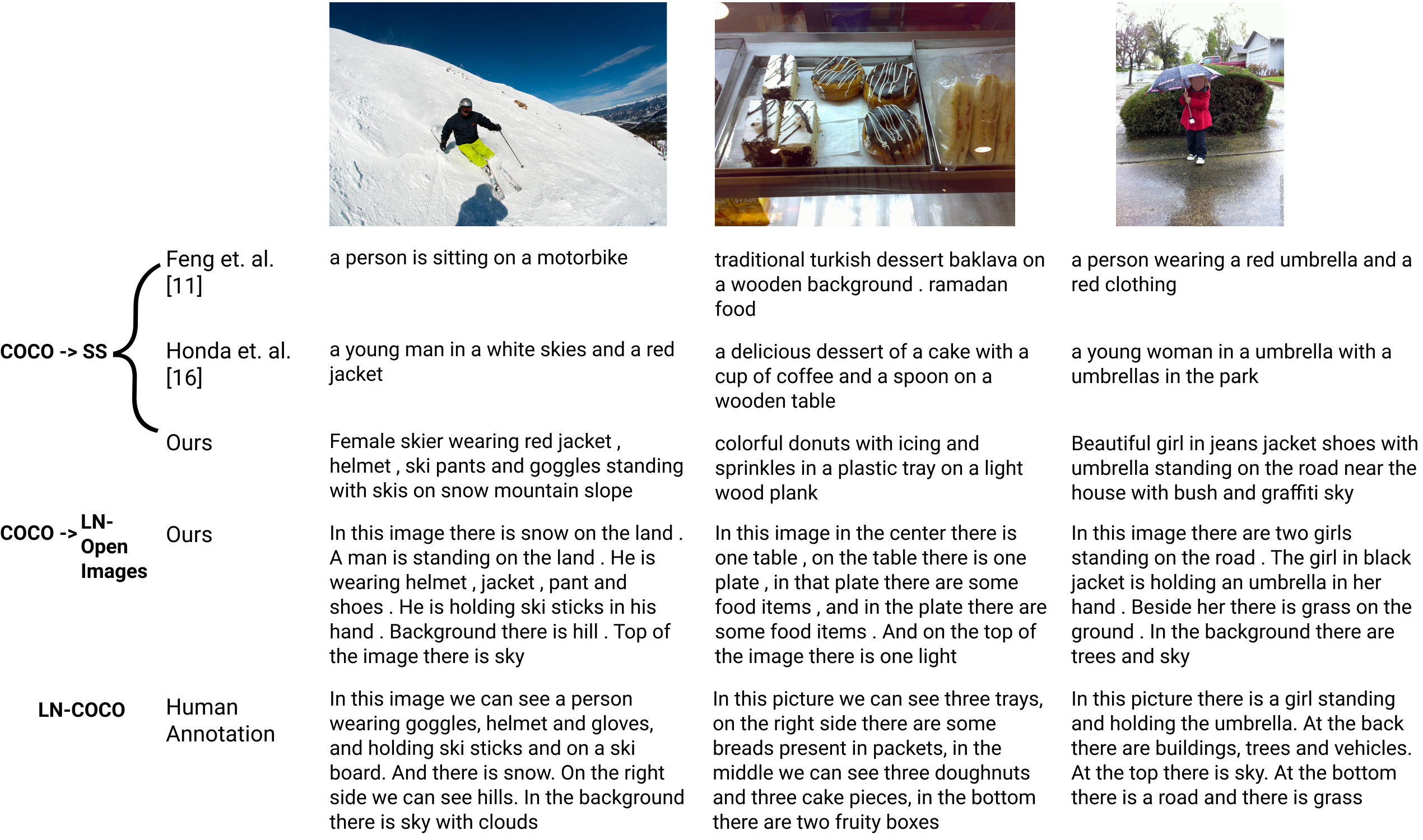}
    \caption{Qualitative results of our method and two baseline methods on COCO test split. 
    }
    \label{fig:coco_ss}
\end{figure*}

Quantitative results are presented in Table \ref{tab:gcc_ss}. We can see that our method outperforms all baseline methods on both text datasets by a clear margin. The qualitative results are in Fig. \ref{fig:coco_ss}. The captions generated by  \cite{feng2019unsupervised} tend to contain verbs describing the objects but make mistakes frequently (e.g., in ``a person is sitting on a motorbike'' the verb ``sitting'' matches with object ``motorbike'' but does not match the input image). \cite{honda2021removing} generated mostly correct captions but the captions are mainly concatenations of object names (nouns) without verbs describing the action of the objects or the relationships between objects (e.g., ``a young man in a white skies''). Our method generates more comprehensive captions with mostly correct nouns and verbs and the captions are aware of the interaction between objects for most part (e.g., ``female skier wearing red jacket'', ``standing with skis on snow mountain slope''). 

We can also see that all methods fail to identify any color attributes, and this is a limitation of only having a pre-trained object detector with no color attribute detector. If a pre-trained color detector becomes available, our method naturally enables the utilization of this information as described in Sec. \ref{sec:utilize_more_info}.


\subsection{Utilizing Object Relationships and Attributes}
\label{sec:exp_rel}

In Sec. \ref{sec:utilize_more_info}, we described how our framework naturally enables utilizing additional pretrained models when they become available. Here, we provide empirical results from adding both a relationship and attribute detector. We use the relationship and attribute detector pretrained on Visual Genome \cite{zellers2018neural} for images, and the semantic parser provided by \cite{anderson2016spice}, which is built on \cite{schuster2015generating}, to find relationship triplets from sentences. Note that we only need the pretrained relationship detectors during training to construct $(R_i,S_i)$ pairs, not test time. We train the model on COCO images, and choose LN-OpenImages as the text dataset since it contains rich semantic relationship information. Ideally, when given a pair of triplets, we would like to completely match the subject-relation-object, but we found that such complete matches are rare because the relationship detector is not 100\% accurate and discrepancies between the image and sentence relationship detectors exist. So, instead, we consider two triplets a match as long as the subject-object matches between them. 

The quantitative results are presented in Table \ref{tab:rel}. Row 2 shows results when a attribute detector is added while row 3 gives the results for when only a relationship detector is added. Results from row 4 come from adding both a relationship and attribute detector. We can see that adding either a relationship or attribute detector is beneficial. Interestingly, row 4 shows a slight improvement over row 3 but a much larger boost over row 2. Qualitatively, we can refer to the last row of Fig. \ref{fig:coco_ss}. We can see that the captions are comprehensive and contain many relationship triplets in the generated captions (e.g., ``he-holding-ski sticks'', ``he-wearing-helmet'', ``man-standing on-land'').

\subsection{Ablations}
\label{sec:ablations}
\subsubsection{Effects of Object Mining}
\label{sec:verify_hyp}

\begin{table*}[!ht]
    \centering
    \resizebox{1.01\columnwidth}{!}{
    \begin{tabular}{l | c c c c c c c }
        \thickhline
        Methods & BLEU-1 & BLEU-4 & METEOR & ROUGE$_L$ & CIDEr & SPICE & WMD \\
        \hline 
        Supervised training  & 0.306 & 0.082 & 0.151 & 0.306 & 0.263 & 0.238 & 0.113 \\
        After replacing objects & 0.297 & 0.078 & 0.145 & 0.298 & 0.227 & 0.232 & 0.109 \\
        After replacing objects and location & 0.298 & 0.080 & 0.146 & 0.300 & 0.239 & 0.234 & 0.108 \\
        Ours & 0.298 & 0.071 & 0.159 & 0.264 & 0.125 & 0.164 & 0.083 \\
        \thickhline
    \end{tabular}
    }
    \caption{The performance of supervised training, before and after replacing the objects with randomly mined objects of the same category (the supervised training was conducted following \cite{meng2021connecting} using their provided code). Detailed interpretation of each row is in Sec. \ref{sec:verify_hyp}}
    \label{tab:verify_hyp1}
\end{table*}

In the set of experiments depicted in Table \ref{tab:verify_hyp1}, we attempt to understand the effect of our proposal to use objects from different images. Our experiments involve a state of the art supervised image captioning method described in \cite{meng2021connecting}, which is a transformer based model. We adopt the same experiment settings but deleted the head for trace in their model, which the authors of \cite{meng2021connecting} utilized in addition to captions and images. We took the code provided, and trained a captioning model with COCO images and LN-COCO. The resulting performance is provided in the {\it first row} of Table \ref{tab:verify_hyp1}. Then, for each image-caption pair in the training set, we detect all object regions in the image and replace each object region with another one of the same category mined from other images in the dataset. During training, we feed the visual features of the substitute objects, together with the 5-D location vectors of the \emph{replaced} objects into the transformer. The performance of this second model is given in the {\it second row} of Table \ref{tab:verify_hyp1}. The third model is trained similarly as the second but with the locations taken from the images where the substitute objects come from. The result is in the {\it third row} of Table \ref{tab:verify_hyp1}. Finally, we run our unsupervised method, which performed as given in the {\it last row} of Table \ref{tab:verify_hyp1}. To better understand the effects of our proposed method, we highlight the difference between these four models in the following.
\begin{enumerate*}[label=(\roman*)]
    \item Row 1, which is the supervised model, is a transformer model as mentioned. Similar to our method, objects are fed into the model, but the key differences are, given an image-caption pair, (1) the objects are all from a single image, and (2) the objects include those that are not in the given caption, which can include background objects.
    \item Row 2 is essentially the same as row 1, but the objects are from different images.
    \item Row 3 is the same as row 2 but the locations are now from the substitute objects' locations in the images they came from. We can think of row 3 as an approximate upper bound to our method.
    \item Our method is the same as row 3 but without the benefit of the additional objects mentioned for row 1.
\end{enumerate*}
We can see that after replacing the object regions with randomly picked objects of the same category, the performance arguably did not drop as much as expected, 
%
which empirically validates our proposal to use object regions from different images to improve object coverage. Interestingly, we can also tell by comparing the second and third rows that the location of the bounding box does not have much influence. From the last row, we can see that our \emph{unsupervised} method performs close to the upper bound (third row) on BLEU, and even outperforms the supervised method on METEOR, while having larger than 20\% relative performance drop on
CIDEr and SPICE. 
One reasonable way to interpret the results is that our model performs relatively better on shorter $n$-grams (e.g., $n=1$), and worse on longer $n$-grams (e.g., $n=4$). In addition, our model performs well when synonyms are considered as correct since METEOR utilizes synonyms while CIDEr requires a stricter match of $n$-grams ($n$ ranges from 1 to 4).

\begin{table*}[!h]
    \centering
    \small
    \setlength{\tabcolsep}{4pt}
    \resizebox{1.01\columnwidth}{!}{
    \begin{tabular}{l l | c c c c c c c c }
        \thickhline
        Text dataset & Method & BLEU-1 & BLEU-4 & METEOR & ROUGE$_L$ & CIDEr & SPICE & WMD & Object Coverage \\
        \hline 
        SS & Ours & \textbf{0.058} & \textbf{0.003} & \textbf{0.060} & \textbf{0.126} & \textbf{0.039} & \textbf{0.102} & \textbf{0.060} & \textbf{100\%}\\
        SS & Ours-half-obj & 0.056 & 0.002 & 0.055 & 0.117 & 0.036 & 0.082 & 0.055 & 50.0\% \\
        SS & Ours-baseline1 & 0.040 & 0.002 & 0.045 & 0.104 & 0.021 & 0.068 & 0.049 & 57.4\% \\
        \hline
        GCC & Ours  & \textbf{0.062} & \textbf{0.004} & \textbf{0.062} & \textbf{0.146} & \textbf{0.032} & \textbf{0.104} & \textbf{0.055} & \textbf{100\%} \\
        GCC & Ours-half-obj & 0.057 & 0.003 & 0.054 & 0.140 & 0.025 & 0.080 & 0.050 & 50.0\% \\
        GCC & Ours-baseline1 & 0.046 & 0.002 & 0.048 & 0.131 & 0.025 & 0.073 & 0.047 & 68.8\% \\
        \thickhline
    \end{tabular}
    }
    \caption{Importance of object coverage. Baseline1 is the same transformer  model as ours trained with pseudo $(I_i,S_i)$ pairs. Object Coverage refers to the percent of objects in a given sentence that are captured by the corresponding image during training, averaged over all used training pairs. Our method has 100\% object coverage by construction.}
    \label{tab:gcc_ss_ablation}
\end{table*}

\begin{table*}[!h]
    \centering
    \small
    \setlength{\tabcolsep}{4pt}
    \resizebox{1.01\columnwidth}{!}{
    \begin{tabular}{c c c c c c c c c }
        \thickhline
         BLEU-1 & BLEU-4 & METEOR & ROUGE$_L$ & CIDEr & SPICE & WMD & Overlap &  Object Coverage \\
        \hline 
         \textbf{0.022} &  \textbf{0.001} & \textbf{0.042} & \textbf{0.134} & \textbf{0.026} & \textbf{0.075} &
         \textbf{0.042} & $\geq 2$ & \textbf{47.5\%}\\
         0.020$_{(\downarrow 9\%)}$ & 0.001 & 0.038$_{(\downarrow 10\%)}$ & 0.123$_{(\downarrow 8\%)}$ & 0.017$_{(\downarrow 35\%)}$ & 0.065$_{(\downarrow 13\%)}$ & 0.042 & $\geq 1$ & 32.6\% \\
         0.020$_{(\downarrow 9\%)}$ & 0.001 & 0.035$_{(\downarrow 17\%)}$ & 0.117$_{(\downarrow 13\%)}$ & 0.015$_{(\downarrow 42\%)}$ & 0.051$_{(\downarrow 32\%)}$ & 0.042 & $\geq 0$ & 4.2\% \\
        \thickhline
    \end{tabular}
    }
    \caption{\color{black}We test the importance of object coverage by doing experiments using \cite{honda2021removing} on COCO images and SS text dataset. We change the object coverage of their method by changing the criterion of selecting training pseudo pairs from existing images and sentences. ``Overlap'' refers to how many objects the image and the sentence in a training pair share in common. 
    See our supplement for the detailed explanation of how we change the object coverage. The number in the parenthesis refers to the relative performance drop compared with the first row.}
    \label{tab:honda_ss_ablation}
\end{table*}

\subsubsection{Object Coverage}
\label{sec:ablations_unsup}

While \cite{honda2021removing} and our method all assume a pretrained object detector, a key difference is that the former mines images while our method mines objects. To demonstrate the benefit of our approach more clearly, we construct a baseline method (``Ours-baseline1'') which shares all experimental settings with our method but instead of constructing $(R_i,S_i)$ pairs, it follows \cite{laina2019towards,honda2021removing} to mine pseudo pairs $(I_i,S_i)$ from the training set. The results are in Table \ref{tab:gcc_ss_ablation}. We observe here that the practice of mining pseudo pairs performs significantly worse than our method, primarily due to the resulting lower object coverage, given as the percentage of object words in the sentence that appear in the corresponding image during training. Our method has a 100\% object coverage by virtue of mining objects, while for Ours-baseline1, there is on average at least a third of the objects that are missing. To further confirm, we artificially drop half of the object coverage from our approach, and the resulting model, denoted as "Ours-half-obj", performs worse than our full approach as expected. Interestingly, Ours-half-obj performs better than Ours-baseline1 even though it has lower object coverage. We further study the importance of object coverage in Table~\ref{tab:honda_ss_ablation}, where we lower the object coverage in the current state of the art method \cite{honda2021removing}. In general, we see that object coverage plays an important role in unsupervised image captioning.


\begin{figure*}[!h]
    \centering
    \includegraphics[width=0.99\linewidth]{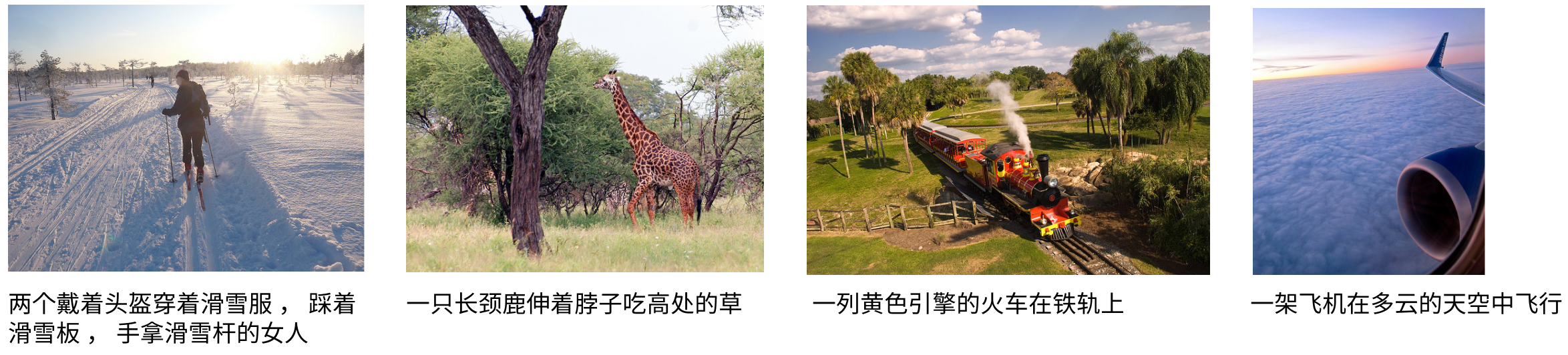}
    \caption{Qualitative results of our method trained on COCO images and COCO-CN (Chinese) captions.}
    \label{fig:coco_chinese}
\end{figure*}

\begin{figure*}[!h]
    \centering
    \includegraphics[width=0.99\linewidth]{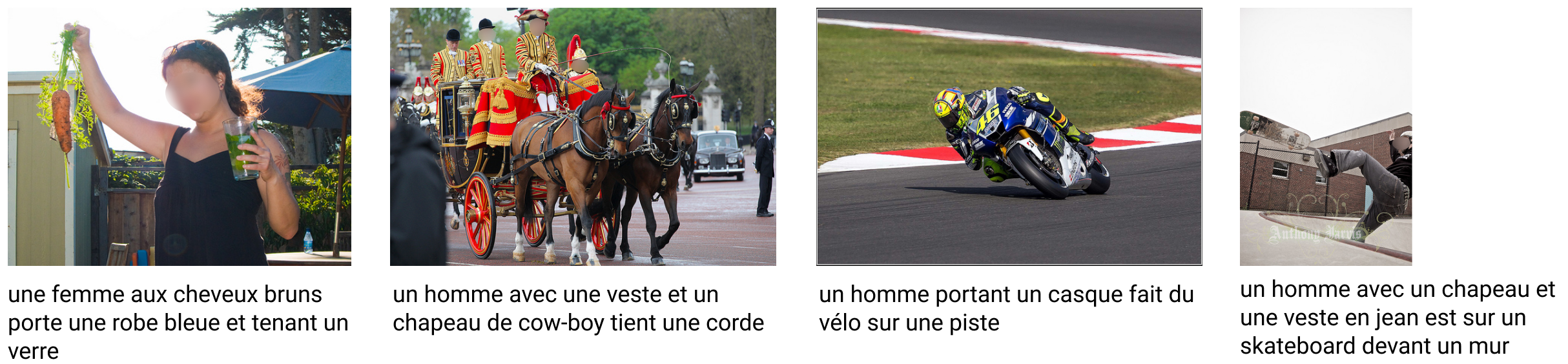}
    \caption{Qualitative results of our method trained on COCO images and Multi30k (French) sentences.}
    \label{fig:coco_french}
\end{figure*}

\begin{table*}[!h]
    \centering
    \small
    \setlength{\tabcolsep}{4pt}
    \resizebox{0.99\textwidth}{!}{
    \begin{tabular}{l l l | c c c c c c c }
        \thickhline
        Training  text & Test text & Method & BLEU-1 & BLEU-4 & METEOR & ROUGE$_L$ & CIDEr & SPICE & WMD \\
        COCO-CN (Chinese) & COCO-CN (Chinese) & Ours & \textbf{0.256} & \textbf{0.039} & 0.121 & 0.228 & \textbf{0.349} & \textbf{0.117} & 0.272 \\
        COCO-CN (Chinese) & COCO-CN (Chinese) & \cite{honda2021removing} & 0.240 & 0.026 & \textbf{0.128} & \textbf{0.234} & 0.197 & 0.110 & \textbf{0.304} \\
        COCO caption (English) & COCO-CN (Chinese) & Baseline-translate & 0.111 & 0.000 & 0.072 & 0.097 & 0.059 & 0.003 & 0.185\\
        COCO-CN (Chinese) & COCO-CN (Chinese) & Baseline-existing & 0.176 & 0.020 & 0.119 & 0.189 & 0.198 & 0.047 & 0.259 \\
        \hline
        Multi30k (French) & Multi30k (French) & Ours & \textbf{0.174} & \textbf{0.010} & \textbf{0.094} & \textbf{0.173} & \textbf{0.120} & \textbf{0.014} & \textbf{0.079} \\
        Multi30k (French) & Multi30k (French) & \cite{honda2021removing}  & 0.143 & 0.007 & 0.083 & 0.156 & 0.053 & 0.014 & 0.072  \\
        Multi30k (English) & Multi30k (French) & Baseline-translate & 0.104 & 0.000 & 0.053 & 0.092 & 0.099 & 0.006 & 0.056\\
        Multi30k (French) & Multi30k (French) & Baseline-existing & 0.126 & 0.008 & 0.086 & 0.154 & 0.059 & 0.013 & 0.078\\
        
        \thickhline
    \end{tabular}
    }
    \caption{Quantitative results of our method on COCO images and two text datasets of different languages (Chinese and French). ``Baseline-translate'' and ``Baseline-existing'' are defined in Sec. \ref{sec:non-eng}.}
    \label{tab:chinese}
\end{table*}

\subsection{Non-English Image Captioning}
\label{sec:non-eng}

Due to the scarcity of annotated image-caption pairs in non-English languages, we expect unsupervised image captioning methods to have a fairly large impact in bridging the gap here. To understand how well our proposed method works for non-English image captioning, we chose Chinese and French to benchmark our method. Our aim is to demonstrate the unpairing that is made possible by our method. To do so, for the experiments in this section, we construct training pairs that are not part of any annotated pairs. 

For Chinese, we use the COCO-CN \cite{li2019coco}, which provides Chinese captions on 20,342 images in COCO. We unpair the training set by training with the train split of COCO-CN captions but in conjunction with the COCO images that are not included in the 20,342 (COCO has a total of $\approx$123k images). Evaluation is then conducted on the test split of COCO-CN captions and the corresponding COCO images. For French, we use Multi30k \cite{elliott2016multi30k} that provides French captions for all of Flickr30k images and 1k images from COCO. Similarly, to achieve the effect of unpairing, the model is trained on just the Flickr30k captions together with the COCO images that are not included in the 1k COCO images. Testing is subsequently conducted on the 1k image-caption COCO pairs.

Three baselines are used here. The first baseline, ``Baseline-translate'', is our method trained to produce English captions, which are then translated. The second baseline, ``Baseline-existing'', follows the same procedure of our method but instead of constructing $(R_i,S_i)$ pairs, it mines pseudo image-caption pairs $(I_s,S_i)$ from existing image and text datasets (similar to the ``baseline1'' in Sec. \ref{sec:ablations_unsup}, but extended to non-English datasets as described in Sec.~\ref{sec:method_non_english}). The last baseline is the current state of the art, \cite{honda2021removing}, which is extended to mine pseudo non-English image-caption pairs (Sec.~\ref{sec:method_non_english}). For all these baselines, we apply the same principle to unpair the training set while ensuring that the training sets for the baselines and our method are apple to apple. Therefore, all the baselines adopt the same training set described in the previous paragraph, except for ``Baseline-translate'', which uses the English version of the same training captions.

We present the quantitative results in Table \ref{tab:chinese}. We can see that our method when trained directly on the target language performs better than ``Baseline-translate'', most likely because translating a full sentence accurately is much harder than translating a single object word. ``Baseline-existing'' performs much better but still lags behind our method as expected based on results presented in the earlier section. {\color{black}\cite{honda2021removing} (with modifications by us extending it to non-English) performs worse than our method on French but has a smaller gap or on par to our method on Chinese. This may be because that the training images and sentences are from the same paired dataset (COCO-CN), thus mining pairs from existing images and sentences are relatively easy. We have also observed in both the English and Chinese datasets that the captions generated by \cite{honda2021removing} usually cover most needed object words although being semantically less meaningful and comprehensive compared with our method, and the captions in COCO-CN are mostly short (like COCO captions), which favors \cite{honda2021removing} when used as the evaluation set.} Qualitative results are also provided  in Fig. \ref{fig:coco_chinese} (Chinese) and Fig. \ref{fig:coco_french} (French). We can see that our method generate reasonable captions in both languages.

\section{Limitations and Discussions}

Unsupervised image captioning is an important research area that has the potential of truly scaling up image captioning in the wild. As progress is made, the hope is that image captioning will soon take a departure from the need to laboriously annotate image-caption pairs, allowing researchers to ``simply'' scrape unpaired images and text from different sources, including the internet. Our work in this paper takes a step towards this goal by showing how one can mine objects from multiple images to improve object coverage. The results we presented in this paper is encouraging, perhaps even surprising that ``unrelated objects'' mined can produce results that surpass current state of the art. By employing a transformer architecture, we also show that our proposed approach is flexible enough to ingest additional information such as object relationships and attributes that we have shown could be immensely useful. The work is not completed yet. Unsupervised performance still lags that of supervised models, but even in that we see some encouraging signs (Table~\ref{tab:verify_hyp1}). Further, the current line of work in unsupervised image captioning including ours depends heavily on a pretrained object detector. 
The class vocabulary size of the pretrained object detector limits the range of images and text we can utilize, since the the objects in the imgages need to be detected by the pretrained object detector, and the class vocabulary of the pretrained object detector may also act as a useful filter when we crawl sentences from the internet.
One way to overcome this limitation is to employ zero-shot object detection that can identify classes not in the training set, especially some classes the practitioner is interested in, but this line of research still has some way to go in terms of maturity. Another more promising direction is for the community to continue to strengthen the performance of supervised object detection as well as broaden the classes of objects covered. Considering that the pretrained object detector has become the 
backbone for various vision tasks, it is expectable that we can see a much stronger object detector in the near future.


\clearpage
\clearpage
{\small
\bibliographystyle{splncs04}
\bibliography{ref}
}

\end{document}